%% file: root.tex
\documentclass[letterpaper, 10 pt, conference]{ieeeconf}  

\IEEEoverridecommandlockouts                              

\usepackage{graphics} 
\usepackage{epsfig} 
\usepackage{mathptmx} 
\usepackage{times} 
\usepackage{amsmath} 
\usepackage{amssymb}  

\usepackage{graphics} 
\usepackage{epsfig} 
\usepackage{amsmath}
\usepackage{amssymb}
\usepackage{multirow}
\usepackage[small]{caption}
\usepackage{xcolor}
\usepackage{hyperref}
\usepackage{array}
\usepackage{booktabs}
\usepackage{makecell}
\usepackage{colortbl}
\usepackage{wrapfig}
\usepackage{graphicx}
\usepackage{balance}
\usepackage{pifont}
\usepackage{subfiles}
\usepackage[noend]{algpseudocode}
\usepackage{algorithm}

\newcolumntype{B}{>{\columncolor[gray]{0.9}}r}
\newcommand{\best}[1]{\textbf{\textcolor[HTML]{990000}{#1}}}

\newcommand{\cmark}{\ding{51}}%
\newcommand{\xmark}{\ding{55}}%
\newcommand{\rgb}{\textbf{\textcolor[HTML]{3BAD6E}{RGB}}}
\newcommand{\rgbs}{\textbf{\textcolor[HTML]{A52A5A}{RGB(S)}}}
\newcommand{\rgbcs}{\textbf{\textcolor[HTML]{A52A5A}{RGB $\times$ Saliency}}}
\newcommand{\rgbso}{\textbf{\textcolor[HTML]{113ED9}{RGB+Saliency (Ours)}}}
\newcommand{\rgbspo}{\textbf{\textcolor[HTML]{EF6682}{RGB+Saliency (PO)}}}
\newcommand{\saliency}{\textbf{\textcolor[HTML]{000000}{Saliency}}}
\newcommand{\rad}{\textbf{\textcolor[HTML]{000000}{RAD}}}
\newcommand{\curl}{\textbf{\textcolor[HTML]{000000}{CURL}}}

\usepackage{titlesec}
\titlespacing*{\subsection}{0pt}{0.15\baselineskip}{0.05\baselineskip}
\titlespacing*{\section}{0pt}{0.3\baselineskip}{0.1\baselineskip}

\title{\LARGE \bf
ViSaRL: Visual Reinforcement Learning Guided by Human Saliency
}

\author{Anthony Liang, Jesse Thomason and Erdem B{\i}y{\i}k
\thanks{All authors are with the Thomas Lord Department of Computer Science, University of Southern California, Los Angeles, CA 90089 USA. 
(Correspondence to: \href{mailto:anthony.liang@usc.edu}{anthony.liang@usc.edu}).}
}

\begin{document}

\maketitle
\thispagestyle{empty}
\pagestyle{empty}

\begin{abstract}
\subfile{sections/abstract}

\end{abstract}

\section{INTRODUCTION}
\subfile{sections/introduction}

\section{RELATED WORK}
\subfile{sections/related_work}

\section{VISUAL SALIENCY-GUIDED RL}
\subfile{sections/visarl}

\section{EXPERIMENT SETUP}
\subfile{sections/experiment_setup}

\section{SIMULATION EXPERIMENTS}
\subfile{sections/sim_exps}

\section{REAL ROBOT EXPERIMENTS}
\subfile{sections/real_exps}

\section{CONCLUSION}
\subfile{sections/conclusion}




\balance
\bibliographystyle{IEEEtran}
\bibliography{references.bib}

\section*{APPENDIX}

\subfile{sections/appendix/environment_description}

\subfile{sections/appendix/hyperparameters}

\subfile{sections/appendix/implementation_details}

\subfile{sections/appendix/real_robot_il}

\subfile{sections/appendix/saliency_predictor}

\end{document}

%% file: sections/abstract.tex
\label{sec:abstract}
Training robots to perform complex control tasks from high-dimensional pixel input using reinforcement learning (RL) is sample-inefficient, because image observations are comprised primarily of task-irrelevant information. 
By contrast, humans are able to visually attend to task-relevant objects and areas.
Based on this insight, we introduce \textbf{Vi}sual \textbf{Sa}liency-Guided \textbf{R}einforcement \textbf{L}earning (ViSaRL).
Using ViSaRL to learn visual representations significantly improves the success rate, sample efficiency, and generalization of an RL agent on diverse tasks including DeepMind Control benchmark, robot manipulation in simulation and on a real robot.
We present approaches for incorporating saliency into both CNN and Transformer-based encoders.
We show that visual representations learned using ViSaRL are robust to various sources of visual perturbations including perceptual noise and scene variations.
ViSaRL nearly doubles success rate on the real-robot tasks compared to the baseline which does not use saliency.

%% file: sections/introduction.tex
\label{sec:introduction} 
Studies in neuroscience \cite{darby2021development} show that humans utilize selective attention to focus on task-relevant information for efficiently processing and understanding complex visual scenes~\cite{itti1998model}.
We employ selective attention when performing everyday pick-and-place tasks to identify the target objects, focus on the grasp points, and execute precise hand-eye coordination. 
We hypothesize that saliency maps capturing human visual attention is a useful signal to process visual observations for AI agents.
In this paper, we investigate whether \textit{human} visual attention helps \textit{agents} perform tasks.

A key ingredient in solving visual control tasks is to learn visual representations that capture useful features of the sensory input to simplify the decision-making process. 
Many works in the deep reinforcement learning (RL) community have proposed to learn such representations through various self-supervised objectives including contrastive learning \cite{laskin2020curl} and data augmentation \cite{laskin2020reinforcement}. 
By contrast, we focus on self-supervision using \emph{saliency} as additional human domain knowledge to inform the representation of task-relevant features in the visual input while filtering out perceptual noise.

\begin{figure}[t]
\centering
\centerline{\includegraphics[width=\linewidth]{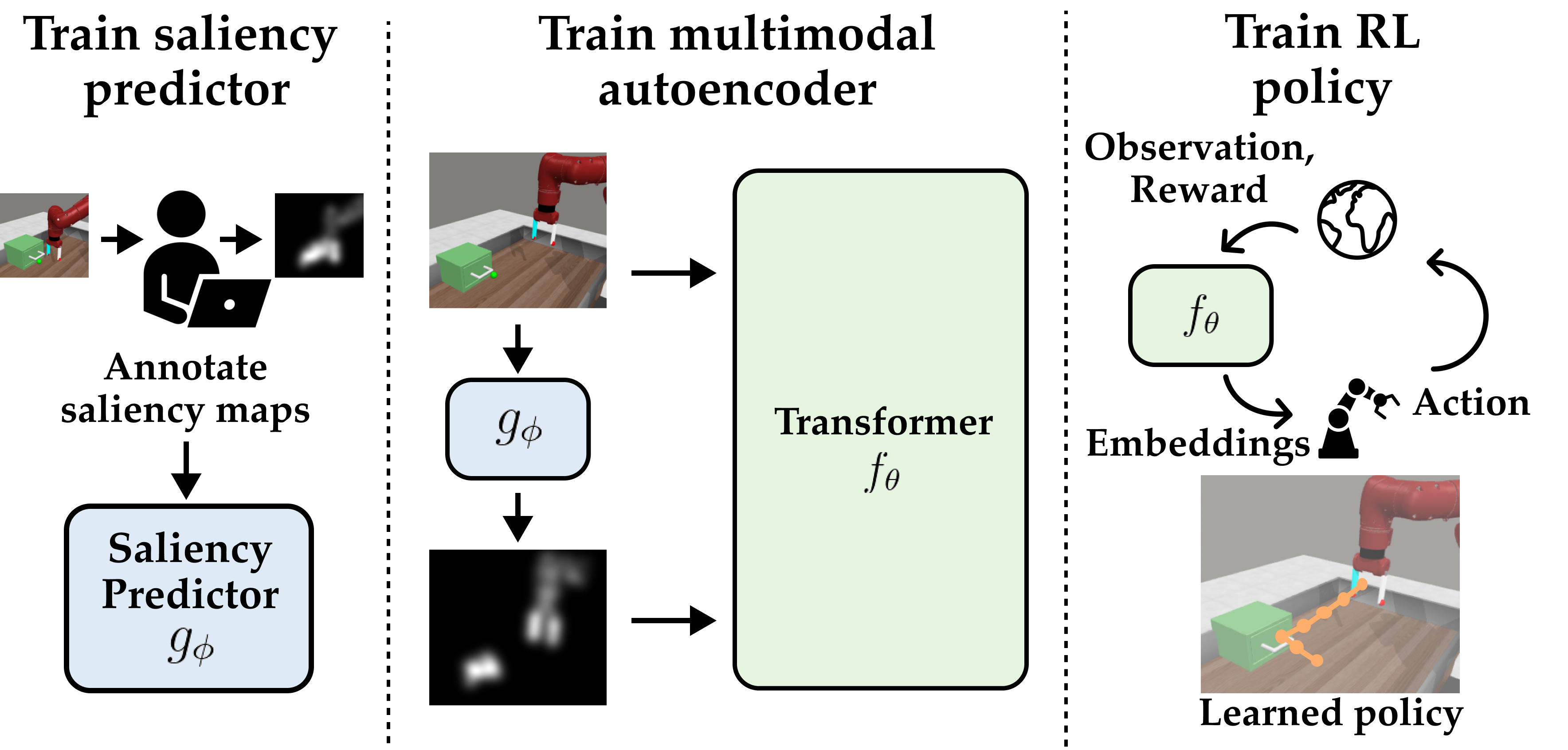}}
\caption{ViSaRL trains a saliency prediction model from a few human-annotated saliency maps. This model is used to augment an offline image dataset with saliency. A visual encoder is pretrained with the dataset and used during downstream policy learning to generate latent representations of the agent's observations.}
\vspace{-5px}
\label{fig:teaser}
\end{figure}

We present \textbf{Vi}sual \textbf{Sa}liency \textbf{R}einforcement \textbf{L}earning (ViSaRL), a general approach for incorporating human-annotated saliency maps as an inductive bias for learned visual representations.
The key idea of ViSaRL is to train a visual encoder using both RGB and saliency inputs and an RL policy that operates over lower dimensional image representations as shown in Figure \ref{fig:teaser}. 
By using a multimodal autoencoder trained using a self-supervised objective, our learned representations attend to the most salient parts of an image for downstream task learning making them robust to visual distractors. 
To circumvent the expensive process of manually annotating saliency maps, we train a state-of-the-art saliency predictor using only a few human-annotated examples to pseudo-label RGB observations with saliency.

We evaluate ViSaRL on a diverse set of challenging continuous control tasks in the DeepMind Control (DMC) suite \cite{tunyasuvunakool2020dm_control} and robot manipulation tasks in Meta-World \cite{yu2020meta} and a real robot. Our method improves in sample-efficiency and robustness over state-of-the-art vision-based RL methods across all environments. Remarkably, ViSaRL nearly doubles the task success rate on a real-robot.

Our contributions can be summarized as follows:
\begin{enumerate}
    \item We propose ViSaRL, a framework for incorporating human-annotated saliency maps to learn robust representations for visual control tasks;
    \item We present approaches for utilizing saliency information in both CNN and Transformer encoders; and
    \item We conduct extensive experiments that demonstrate ViSaRL consistently outperform prior state-of-the-art methods for various visual control tasks both in simulation and on a real robot. 
\end{enumerate}

%% file: sections/related_work.tex
\label{sec:related_work}

Different forms of human data can be leveraged when solving control tasks. 
Researchers have created various interfaces to collect different data modalities from humans such as reward sketches~\cite{cabi2019scaling}, feature traces~\cite{bobu2021feature}, scaled comparisons~\cite{wilde2022learning}, and abstract trajectories~\cite{tao2022executable}.
Attention saliency maps, in contrast, do not require humans to work with abstract concepts like rewards and task features, and do not require watching and comparing lengthy trajectories.

\textbf{Saliency Maps}.
Saliency maps approximate which parts of an image tend to attract human visual attention, corresponding to where the human eye would likely fixate when viewing an image~\cite{tong2010full}. 
Saliency maps have been used in both computer vision and machine learning for various applications including activity recognition~\cite{wang2016beyond}, question answering~\cite{das2017human}, and object segmentation~\cite{li2011saliency}. 
The explainable AI community uses saliency maps to understand how a model is making its predictions and to identify the most informative regions of an image for a particular task~\cite{simonyan2013deep, mundhenk2019efficient, zhao2016person}.
Most existing works explore using saliency maps only as tools for interpretation~\cite{atrey2019exploratory, rosynski2020gradient}. 
For example, Atrey et al.~\cite{atrey2019exploratory} and Rosynski et al.~\cite{rosynski2020gradient} use saliency maps to rationalize and explain the actions of RL agents in Atari games. 
Boyd et al.~\cite{boyd2023cyborg} show saliency maps encoding prior human knowledge enable better generalization of deep learning models.

Bertoin et al.~\cite{bertoin2022look} uses neural network saliency in a self-supervised regularization objective to encourage better visual representations.
We do not use a model's saliency, but rather human saliency to identify salient regions of the input image and distill this knowledge into the visual representation.

\textbf{User Interfaces for Human Saliency}.
ViSaRL needs a small number of human-annotated saliency maps to bootstrap the saliency prediction network.
Prior work used superpixel segmentation~\cite{achanta2012slic} to first divide each image into segments, and then asked humans to click on the segments that are salient~\cite{zhao2016person}. 
However, that method requires manually checking and combining the segments that belong to the same object before showing the images to annotators, burdening system designers. 
As an alternative, Boyd et al.~\cite{boyd2022human} used interfaces where the annotators created binary masks by simply clicking on images. 
We employ a similar but simpler interface: an annotator clicks on the salient parts of the image, and a Gaussian kernel is applied around selected pixels to achieve smooth saliency maps shown in Figure \ref{fig:gui}.

\textbf{Representation Learning for RL}.
Saliency maps are representations of the environment that carry domain knowledge about which regions of the visual input are important for the downstream task. 
Such representations are crucial in RL because they enable agents to tractably deal with high-dimensional image observation spaces.

Prior works have shown self-supervised learning with data augmentation helps achieve good performance in image-based RL. 
Contrastive Unsupervised RL (CURL) \cite{laskin2020curl} employs a contrastive learning objective as an auxiliary loss to learn representations for off-policy RL. 
RL with Augmented Data (RAD)~\cite{laskin2020reinforcement} use simple image augmentations such as random cropping and color jittering as regularization to learn representations invariant to visual perturbations.
ViSaRL does not use data augmentation directly in the value function or policy update. 
Instead, saliency augmentation is introduced during the visual encoder pretraining phase.

Nair et al.~\cite{nair2022r3m} and Karamcheti et al.~\cite{karamcheti2023language} propose to combine internet scale language and vision datasets to learn visual representations applicable across all robot tasks. 
While they focus on learning general visual representations, ViSaRL augments small task-specific datasets with saliency information to improve pretrained visual representations.

Sax et al. \cite{sax2018mid} demonstrated that mid-level visual representations such as surface normals or depth predictions from RGB images removes unimportant information and captures useful invariances about the visual world leading to better success on downstream RL tasks. Similar to Sax et al.~\cite{sax2018mid}, ViSaRL utilizes saliency maps as a mid-level feature. However, we empirically show that our approach for incorporating the saliency information into the visual representation improves task performance over other mid-level features including depth and surface normals.

%% file: sections/visarl.tex
\label{sec:visarl}
\begin{figure}[t]
\begin{center}
\vspace*{0.25cm}
\includegraphics[width=0.6\linewidth]{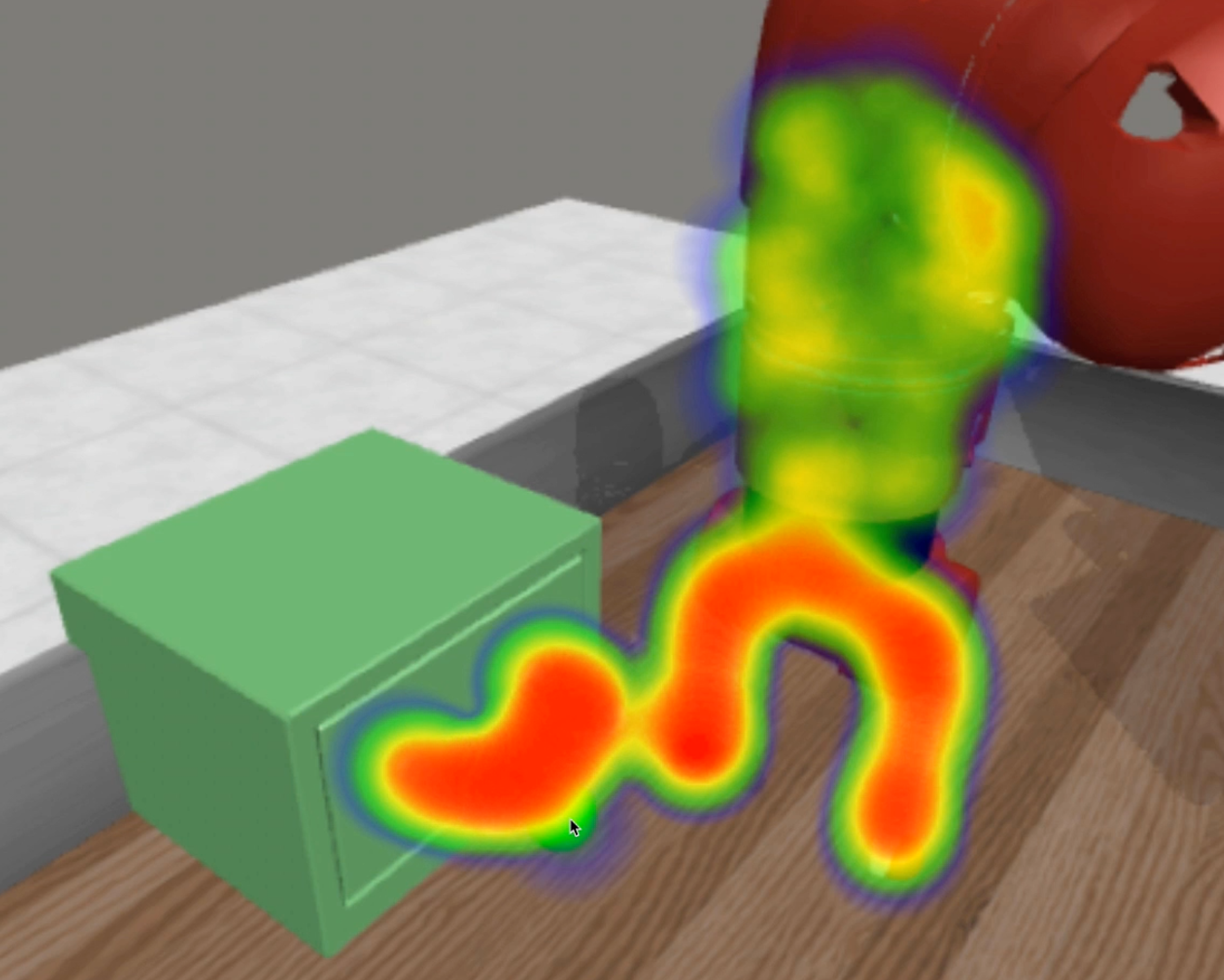}
\caption{\textbf{Annotation Interface}. Custom click-based saliency annotation interface. Each click generates a Gaussian centered at the clicked coordinate with some variance. Warmer colors denote more salient regions such the drawer handle and the robot's end-effector.}
\vspace{-15px}
\label{fig:gui}
\end{center}
\end{figure}

\begin{figure*}[t]
\begin{center}
\centerline{\includegraphics[width=\textwidth]{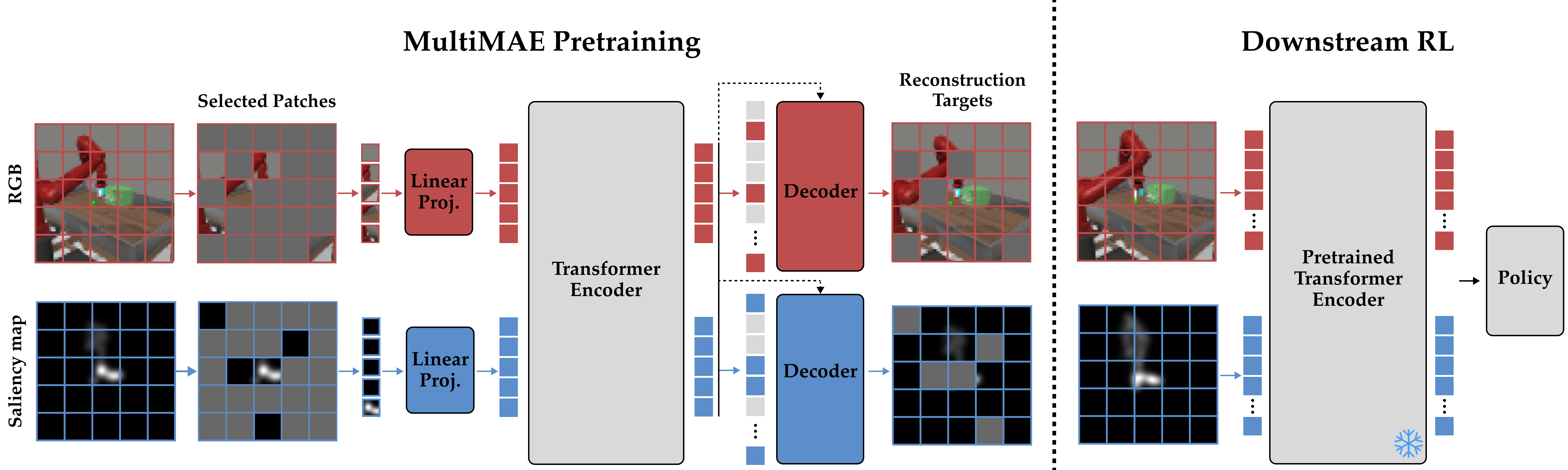}}
\caption{\textbf{ViSaRL}. We pretrain a MultiMAE \cite{bachmann2022multimae} Transformer on a dataset of paired images and saliency maps. MultiMAE employs a self-supervised objective in which masked patches for both input modalities are reconstructed given only the visible patches. The pretrained model is frozen and used for extracting representations during task learning. There is no input masking during downstream RL.}
\vspace{-25px}
\label{fig:model}
\end{center}
\end{figure*}

We propose ViSaRL, a simple approach for incorporating human-annotated saliency to learn representations for visual control tasks.
ViSaRL can be implemented on top of any standard RL algorithm for learning a policy. 
It aims to learn representations that encode useful task-specific inductive biases from human saliency maps.
ViSaRL consists of three learned components: a saliency predictor $g_{\phi}$, an image encoder $f_{\theta}$, and a policy network $\pi_{\psi}$ shown in 
Figure \ref{fig:teaser}. 
We will elaborate on each component in the following sections.


\textbf{Saliency Predictor}. 
Saliency maps highlight regions in an image likely to capture human attention or are considered crucial for a given task. 
Having a human expert annotate saliency maps for every image observation is impractical and not scalable to complex domains. 
To alleviate the burden of manual annotations, we propose to learn a saliency network using only a few hand-annotated examples of saliency maps collected using a custom user interface.

Formally, given an input RGB image observation, $I \in \mathbb{R}^{H \times W \times C}$, a saliency predictor $g_\phi$ maps an input image $I$ to a continuous saliency map ${M=g_\phi(I)\in[0,1]^{H \times W}}$ highlighting important parts of the image for the downstream task. 
We use a state-of-the-art saliency model, Pixel-wise Contextual Attention network (PiCANet) \cite{liu2018picanet}.
PiCANet uses global and local pixel-wise attention modules to selectively attend to informative context. 
Global attention can attend to backgrounds for foreground objects while local attention can attend to regions that have similar appearance.
The mixture of attention at different scales allows for more homogeneous and consistent saliency predictions. 
We emphasize that our method is agnostic to the choice of saliency model.



\textbf{Pretraining Visual Representation}. 
We use our trained $g_{\phi}$ to pseudo-label an offline image dataset collected using any behavior policy (random, replay buffer, expert demonstrations, etc.) with saliency maps.
We then use the paired image and saliency dataset to pretrain an image encoder, $f_{\theta}$.
We experiment with two models for our backbone visual encoder, CNN and Transformer, and investigate different techniques for augmenting each with saliency input.
To add saliency to a CNN, we can use saliency as a continuous mask or simply add it as an additional channel per pixel. 
For a Transformer encoder, we pretrain the model with saliency as an additional input using a masked reconstruction objective.

Masked autoencoders (MAE) \cite{he2022masked} are an effective and scalable approach for learning visual representations. 
MAE masks out random patches of an image and reconstructs the masked patches using a Vision Transformer (ViT) \cite{dosovitskiy2020image}.
An image $I \in \mathbb{R}^{H \times W \times C}$ is processed into a sequence of 2D patches $h \in \mathbb{R}^{K \times (P^{2} C)}$ where $P$ is the patch size and $K = HW / P^{2}$ is the number of patches. 
A subset of these patches are randomly masked out with a masking ratio of $m$. 
Only the \textit{visible, unmasked patches} are used as input to the ViT encoder.
Masking reduces the input sequence length and encourages learning global, contextualized representations.

The image patches are embedded via a linear projection and added to positional embeddings.
The resulting tokens are processed via a series of Transformers.
Finally, a ViT decoder reconstructs the original input by processing all of the tokens including the encoded visible patches and placeholder mask tokens.
Following He et al.~\cite{he2022masked}, we set a high masking ratio $m\!=\!0.75$ and a heavy-encoder, light-decoder architecture.

\begin{algorithm}[t]
    \caption{Visual Saliency-Guided RL}
    \label{alg:visarl}
    \begin{algorithmic}[1]
        \State {\bfseries Input:} env, $\phi, \psi, \theta$ randomly initialized parameters 
        \State Collect image dataset $\mathcal{D}$ with any behavioral policy $\pi_{B}$
        \State Annotate $N$ random frames from $\mathcal{D}$ with saliency
        \State Train $g_{\phi}$ on $\{(I, M)\}_{i=1}^{N}$ using PiCANet loss
        \State Annotate the full dataset $\mathcal{D} = \{(I, g_{\phi}(I))\}_{i=1}^{N}$ 
        \State Train $f_{\theta}$ using masked reconstruction 
        \For{every environment step} \Comment{RL Training}
            \State Select action $a \sim \pi_{\psi}(f_{\theta}(o, g_{\phi}(o)))$
            \State Optimize $\mathcal{L}_{RL}$ with respect to $\psi$
        \EndFor
    \end{algorithmic}
\end{algorithm}

\textbf{MultiMAE for Encoding Saliency.} 
The standard MAE architecture is limited to processing just RGB modality. 
We propose to incorporate saliency using the MultiMAE \cite{bachmann2022multimae} architecture shown in Figure \ref{fig:model}. 
MultiMAE extends MAE to encode multiple input modalities in a way that these modalities are contributing synergistically to the resulting representation.
Specifically, MultiMAE uses a different linear projection and decoder for each input modality.
A cross attention layer is used in each decoder to incorporate information from the encoded tokens of other modalities.
Crucially, MultiMAE's pretraining objective requires the model to perform well in both the original MAE objective of RGB in-painting and cross-modal reconstruction, resulting in a stronger cross-modal visual representation.

\begin{figure*}[t]
\centering
\centerline{\includegraphics[width=\textwidth]{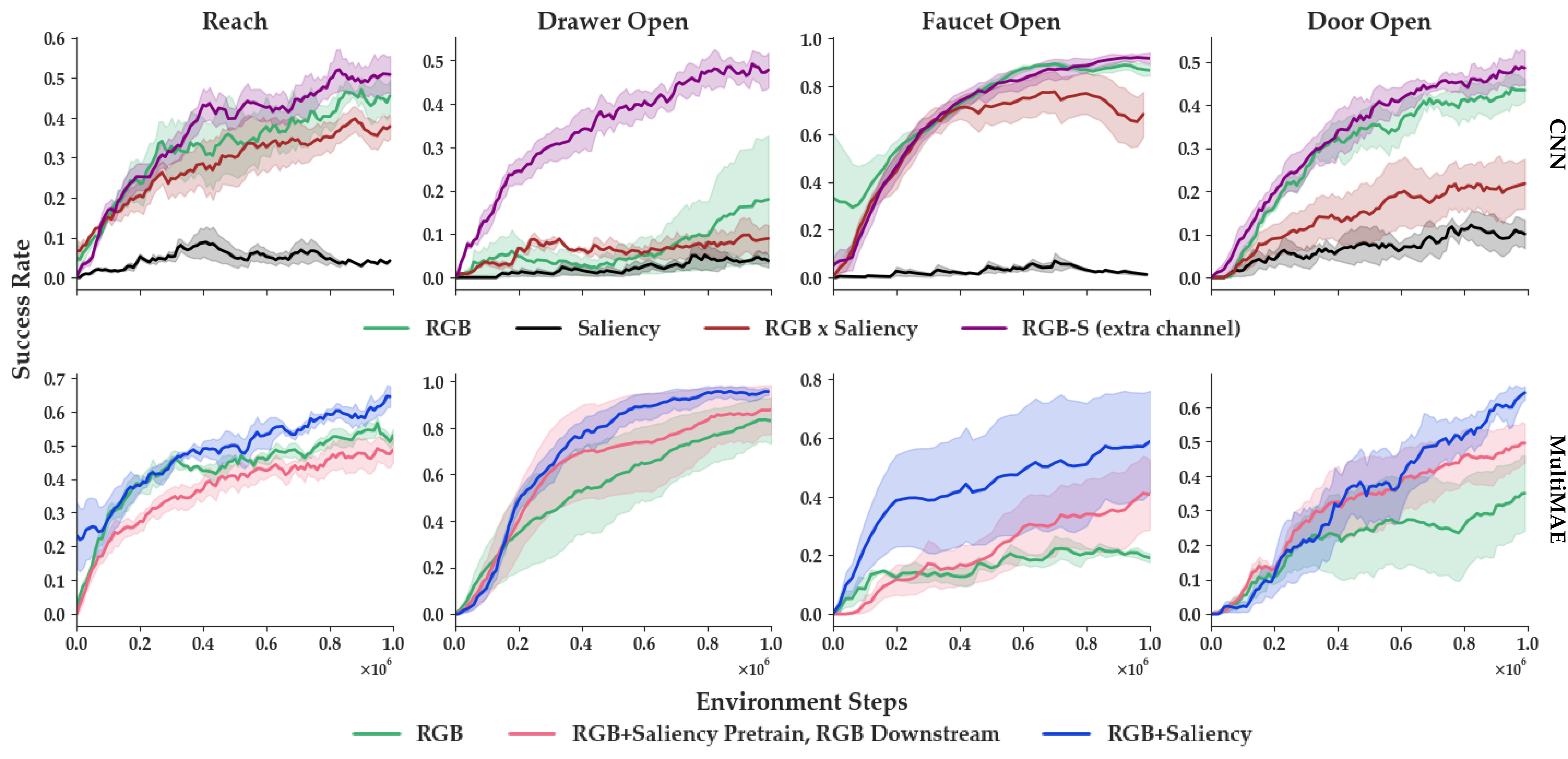}}
\vspace{-5px}
\caption{\textbf{Learning curves} for four robot manipulation tasks in Meta-World evaluated by task success rate. \textbf{(Top)} CNN encoder methods. \textbf{(Bottom)} Transformer encoder methods. 
We select tasks that require manipulating small objects with different motions such as a pushing, pulling, and reaching. 
The solid lines represent the mean and shaded region the standard error across three seeds.}
\vspace{-15px}
\label{fig:result_plots}    
\end{figure*}

\textbf{Downstream Policy Learning.}
After pretraining the MultiMAE model, we freeze the encoder and use it to compute latent representations of environment observations for policy training. 
ViSaRL is not only compatible with online RL algorithms such as Soft-Actor Critic (SAC) \cite{haarnoja2018soft} in which the agent learns through environment interactions but also imitation learning from expert demonstrations. 
Image inputs are not masked during policy learning.
We average the patch embeddings to generate a global image representation. 
The full procedure for ViSaRL is summarized in Algorithm \ref{alg:visarl}.

%% file: sections/experiment_setup.tex
\label{sec:experiment}
To demonstrate the effectiveness of using human-annotated saliency information to enhance visual representations for task learning, we show quantitative results of our approach with two different encoder backbones, CNN and Transformer, across multiple simulated environments including the Meta-World manipulation \cite{yu2020meta} and DMC benchmarks \cite{bertoin2022look} and real-robot manipulation with a Kinova Jaco 2 arm.
We train the downstream policy using SAC \cite{haarnoja2018soft} for the simulation experiments and behavioral cloning with expert demonstrations for the real robot experiments.
\footnote{The code implementation for reproducing the results and additional analysis can be found on: \href{https://liralab.usc.edu/visarl/}{https://liralab.usc.edu/visarl/}.} 


\textbf{Saliency Map Annotation}.
We created a simple user interface to collect saliency annotations shown in Figure~\ref{fig:gui}.
An annotator clicks on the pixels in the image that they think are relevant for performing the given downstream task.
The interface creates a Gaussian centered around the clicked pixel with $\sigma = 10$ on an input image of resolution $256 \times 256 \times 3$. 

\begin{table*}[t]
\centering
\vspace*{0.25cm}
\setlength{\aboverulesep}{0pt}
\begin{tabular}{lcccc|ccc}
\multirow{2}{*}{Meta-World} &\multicolumn{4}{c|}{CNN} & \multicolumn{3}{c}{MMAE} \\
\cmidrule(l){2-8}
 & \rgb & \saliency & \rgbcs & \rgbs & \rgb & \rgbspo & \rgbso \\
\midrule
\texttt{Reach} & $0.40${\scriptsize $\pm 0.12$} & $0.04${\scriptsize $\pm 0.02$} & $0.38${\scriptsize $\pm 0.05$} & $0.52${\scriptsize $\pm 0.08$} & $0.50${\scriptsize $\pm 0.02$} & $0.48${\scriptsize $\pm 0.06$} & \best{$\pmb{0.62}${\scriptsize $\pm 0.06$}} \\
\texttt{Drawer Open} & $0.18${\scriptsize $\pm 0.25$} & $0.04${\scriptsize $\pm 0.02$} & $0.10${\scriptsize $\pm 0.04$} & $0.48${\scriptsize $\pm 0.06$} & $0.84${\scriptsize $\pm 0.02$} & $0.88${\scriptsize $\pm 0.04$} & \best{$\pmb{0.94}${\scriptsize $\pm 0.04$}} \\
\texttt{Faucet Open} & $0.82${\scriptsize $\pm 0.02$} & $0.02${\scriptsize $\pm 0.02$} & $0.72${\scriptsize $\pm 0.16$} & \best{$\pmb{0.86}${\scriptsize $\pm 0.02$}} & $0.18${\scriptsize $\pm 0.05$} & $0.40${\scriptsize $\pm 0.20$} & $0.62${\scriptsize $\pm 0.16$} \\
\texttt{Door Open} & $0.42${\scriptsize $\pm 0.04$} & $0.10${\scriptsize $\pm 0.06$} & $0.22${\scriptsize $\pm 0.10$} & $0.48${\scriptsize $\pm 0.06$} & $0.36${\scriptsize $\pm 0.18$} & $0.52${\scriptsize $\pm 0.08$} & \best{$\pmb{0.64}${\scriptsize $\pm 0.02$}} \\
\midrule
Average & $0.46${\scriptsize $\pm 0.11$} & $0.05${\scriptsize $\pm 0.03$} & $0.36${\scriptsize $\pm 0.08$} & $0.58${\scriptsize $\pm 0.05$} & $0.48${\scriptsize $\pm 0.12$} & $0.57${\scriptsize $\pm 0.10$} & \best{$\pmb{0.65}${\scriptsize $\pm 0.07$}} \\
 \bottomrule
\end{tabular}
\caption{
    \textbf{Success rate} on four Meta-World manipulation tasks averaged across 50 rollouts and 3 seeds for the CNN and MultiMAE (MMAE) visual encoder backbones. Text in \best{maroon} indicates the best performing method per task.}
    \vspace{-20px}
\label{tab:results_mw}
\end{table*}

%% file: sections/sim_exps.tex
\label{sec:sim_exps}
Figure \ref{fig:result_plots} and Table \ref{tab:results_mw} 
summarize our main findings on 4 Meta-World robot manipulation tasks and 5 DMC tasks using CNN and Transformer backbones. 
We compare against two state-of-the-art methods for visual representation learning: CURL~\cite{laskin2020curl},
a contrastive representation learning method and RAD~\cite{laskin2020reinforcement},
a method to combine various image augmentations to induce visual invariances in the learned representations.

\textbf{Saliency input improves downstream task success rates.}~Incorporating saliency improves the task success rate in Meta-World using CNN and Transformer encoders by 13\% and 18\% respectively over the next best baseline. 
For DMC environments, we observe a 256\% relative improvement in average return when using saliency input. 
Our Transformer encoder results in an average 4\% relative improvement in environment returns across all tasks over the next best baseline with a 7.5\% improvement in \texttt{Cartpole Swing}.

\subsection{CNN Encoder}
We follow the CNN implementation used in prior work \cite{laskin2020reinforcement,bertoin2022look} and compare several methods for incorporating saliency. 
In each approach, the CNN encoder and policy are trained jointly but take different inputs.

\textbf{A saliency channel achieves the best task success rate for CNN encoder.} In Table \ref{tab:results_mw}, we find that naive ways of utilizing saliency, such as using saliency directly as input the policy (\saliency), are unable to achieve good performance on the task. 
We hypothesize that the saliency map alone is not sufficient to infer the exact orientation of the end-effector position critical for fine control. 
Supporting this hypothesis, we find that using saliency to mask the RGB observation (\rgbcs) achieves higher task success rate than \textbf{Saliency}, but is still worse than providing the raw RGB input (\rgb). 
Although masking should help the encoder identify the important image features, it may still be nontrivial for the encoder to differentiate between similarly masked observations. 
Lastly, we find that incorporating saliency as an additional channel to the RGB input (\rgbs) improves task success rate by more than $10\%$ across all tasks. 
We hypothesize that the CNN encoder is able to utilize the saliency information to more effectively associate the observed rewards to the relevant features in the image. 

\begin{table}[t]
\centering
\setlength{\aboverulesep}{0pt}
\setlength{\belowrulesep}{0pt}
\begin{tabular}{lccccc}
& DMC-GB & \curl & \rad & \thead{\textbf{\textcolor[HTML]{113ED9}{RGB}} \\ \textbf{\textcolor[HTML]{113ED9}{+Saliency (Ours)}}} \\
\toprule
\multirow{4}{*}{\rotatebox[origin=c]{90}{\texttt{color}}} & \texttt{Walker Walk} & $645${\scriptsize $\pm 55$} & $636${\scriptsize $\pm 33$} & \best{$\pmb{823}${\scriptsize $\pm 55$}} \\
& \texttt{Cartpole Swing} & $668${\scriptsize $\pm 74$} & $763${\scriptsize $\pm 29$} & \best{$\pmb{870}${\scriptsize $\pm 21$}} \\
& \texttt{Ball Catch} & $565${\scriptsize $\pm 160$} & $727${\scriptsize $\pm 87$} & \best{$\pmb{962}${\scriptsize $\pm 14$}} \\
& \texttt{Finger Spin} & $781${\scriptsize $\pm 139$} & $789${\scriptsize $\pm 160$} & \best{$\pmb{823}${\scriptsize $\pm 102$}} \\
\midrule
\multirow{4}{*}{\rotatebox[origin=c]{90}{\texttt{video}}} & \texttt{Walker Walk} & $572${\scriptsize $\pm 121$} & $595${\scriptsize $\pm 85$} & \best{$\pmb{756}${\scriptsize $\pm 42$}} \\
& \texttt{Cartpole Swing} & $418${\scriptsize $\pm 72$} & $434${\scriptsize $\pm 58$} & \best{$\pmb{730}${\scriptsize $\pm 32$}} \\
& \texttt{Ball Catch} & $402${\scriptsize $\pm 169$} & $520${\scriptsize $\pm 44$} & \best{$\pmb{802}${\scriptsize $\pm 78$}} \\
& \texttt{Finger Spin} & $612${\scriptsize $\pm 55$} & $588${\scriptsize $\pm 82$} & \best{$\pmb{702}${\scriptsize $\pm 83$}} \\
\bottomrule
\end{tabular}
\caption{
    \textbf{Average return} of ViSaRL and baseline methods on the \texttt{color} and \texttt{video} environments from DMC-GB.
}
\vspace{-5px}
\label{tab:results_dmc_gb}
\end{table}

\subsection{MultiMAE Transformer}

We compare MultiMAE representations pretrained with RGB only (\rgb) and both RGB and saliency (\rgbspo, \rgbso). \rgbspo \ uses saliency only during pretraining while \rgbso \ uses saliency in both pretraining and downstream RL.

\textbf{Training encoder with saliency improves RGB-only success rates at inference time.} Even without saliency input during downstream RL, using saliency as an additional input modality during pretraining still improves downstream performance on 3 of the 4 tasks. Except for the \texttt{Reach} task, where performances are similar, \rgbspo \ achieves better success rate than \rgb, with an average absolute gain of 10\% across tasks.

\textbf{Using saliency in both pretraining and inference yields the best performance.} 
We compare the full ViSaRL method (\rgbso) to pretraining using only the RGB images (\rgb) in Tables \ref{tab:results_mw} 
demonstrating that multimodal pretraining with saliency information significantly outperforms single modality pretraining by at least a 10\% margin across all tasks. 
Notably, \rgb \ achieves only 19\% success on \texttt{Faucet Open}, while our approach solves the task with 62\% success rate. 
Using saliency as an input for both pretraining and downstream RL (\rgbso) improves task success rate over \rgbspo \ because there are new observations during online training that were not in the pretraining dataset. 

\textbf{ViSaRL representations generalize to unseen environments.} 
We evaluate the generalizability of our learned representations on the challenging \textit{random colors} and \textit{video backgrounds} benchmark from DMControl-GB \cite{hansen2021generalization}. 
In DMControl-GB, agents trained in the original environment are evaluated on their generalization to the same environment with visually perturbed backgrounds using randomized color and video overlays.
ViSaRL significantly outperforms the baselines across all tasks as shown in Table \ref{tab:results_dmc_gb}, with an average 19\% and 35\% relative improvement respectively for the \texttt{color} and \texttt{video} settings. 

\textbf{Human-annotated saliency improves performance compared to depth and surface normals.} 
We conduct ablation experiments to compare saliency versus other mid-level input modalities such as depth and surface normals proposed by Sax et al.~\cite{sax2018mid}. 
We substitute saliency with these other modalities as input to the MultiMAE. 
In Table \ref{tab:ablations}, we observe that neither depth nor surface normal features alone improves task success over just using RGB image input. 
By contrast, adding saliency as an additional modality consistently improves task success suggesting that human-annotated saliency information can help learn better visual representations compared to other input modalities.

\begin{table}[t]
\centering
\setlength{\aboverulesep}{0pt}
\setlength{\belowrulesep}{0pt}
\begin{tabular}{lcccc}
& \thead{Saliency} & \thead{\rgb}     & \thead{RGB + Depth} & \thead{RGB + SN}         \\
 \toprule
\multirow{2}{*}{\texttt{Reach}} & \xmark & $0.50${\scriptsize $\pm0.02$} & $0.43${\scriptsize $\pm0.07$} & $0.46${\scriptsize $\pm0.04$} \\
                       & \cmark & \best{$\pmb{0.62}${\scriptsize $\pm0.06$}} & \best{$\pmb{0.58}${\scriptsize $\pm0.04$}} & \best{$\pmb{0.64}${\scriptsize $\pm0.06$}} \\
 \midrule
\multirow{2}{*}{\texttt{\shortstack[l]{Drawer \\ Open}}} & \xmark & $0.82${\scriptsize $\pm0.02$} & $0.76${\scriptsize $\pm0.06$} & $0.80${\scriptsize $\pm0.04$} \\
                                               & \cmark & \best{$\pmb{0.94}${\scriptsize $\pm0.04$}} & \best{$\pmb{0.90}${\scriptsize $\pm0.04$}} & \best{$\pmb{0.92}${\scriptsize $\pm0.04$}} \\
\midrule
\multirow{2}{*}{\texttt{\shortstack[l]{Faucet \\ Open}}} & \xmark & $0.18${\scriptsize $\pm0.04$} & $0.22${\scriptsize $\pm0.04$} & $0.24${\scriptsize $\pm0.04$} \\
                                               & \cmark & \best{$\pmb{0.62}${\scriptsize $\pm0.16$}} & \best{$\pmb{0.54}${\scriptsize $\pm0.06$}} & \best{$\pmb{0.58}${\scriptsize $\pm0.10$}} \\
\midrule
\multirow{2}{*}{\texttt{\shortstack[l]{Door \\ Open}}} & \xmark & $0.36${\scriptsize $\pm0.18$} & $0.28${\scriptsize $\pm0.14$} & $0.34${\scriptsize $\pm0.10$} \\
                                               & \cmark & \best{$\pmb{0.64}${\scriptsize $\pm0.02$}} & \best{$\pmb{0.62}${\scriptsize $\pm0.04$}} & \best{$\pmb{0.58}${\scriptsize $\pm0.04$}} \\
 \bottomrule
\end{tabular}
\caption{
    Human-annotated saliency versus depth and surface normals (SN) as input modalities to MultiMAE model.
} 
\label{tab:ablations}
\end{table}

%% file: sections/real_exps.tex
\begin{figure}[t]
\begin{center}
\includegraphics[width=\linewidth]{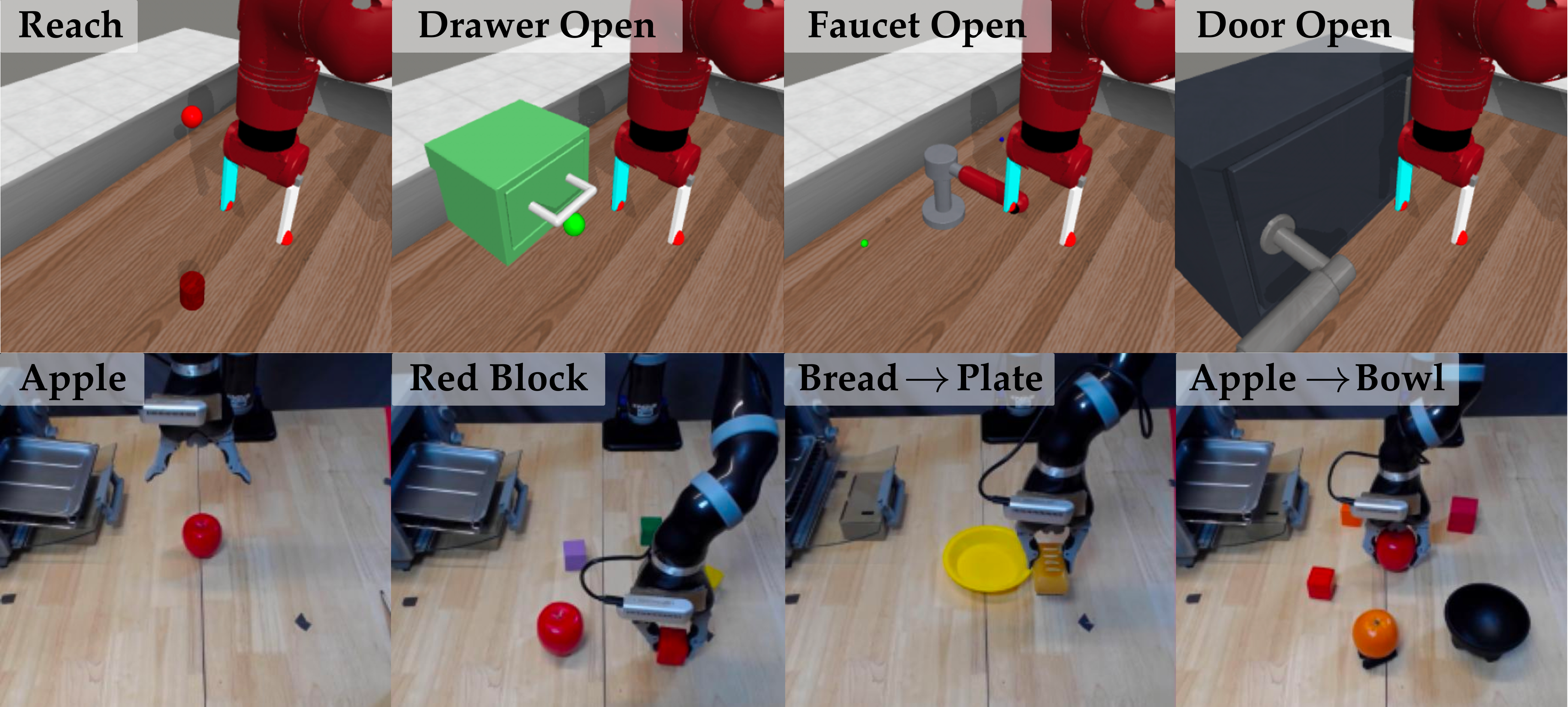}
\vspace{-15px}
\caption{\textbf{Evaluation Tasks}. Four Meta-World (top) simulation tasks and four real-robot tabletop manipulation tasks (bottom).}
\vspace{-15px}
\label{fig:tasks}
\end{center}
\end{figure}

We use a Kinova Jaco 2 (6-DoF) robot arm with a 1-DoF gripper. 
The observation space consists of a front-view image ($224 \times 224 \times 3$) from a Logitech webcam and proprioceptive information. 
We consider four tabletop manipulation tasks shown in Figure \ref{fig:tasks}. 
In two of these tasks, we purposefully include distractor objects to evaluate the robustness of our learned representations to scene variations.
We collected 10 demonstrations per task, resulting in an offline imitation learning dataset of around 10,000 transitions.

For each task, 10 randomly sampled frames are hand-annotated with saliency. 
Even with real-world images, only a small number of annotated frames are required to learn a good saliency predictor.
We train an imitation learning policy by minimizing the mean-squared error between predicted end-effector pose and expert actions. 
We use a recurrent policy to encode history information and a 2-layer MLP to predict continuous actions.
\begin{table}[t]
\centering
\setlength{\aboverulesep}{0pt}
\setlength{\belowrulesep}{0pt}
\begin{tabular}{lrrrrB}
MultiMAE                     & Apple & \makecell{Red \\ Block} & \makecell{Bread \\ $\rightarrow$ Plate} & \makecell{Apple \\ $\rightarrow$ Bowl} 
& Cumulative \\ 
\toprule
\textbf{\textcolor[HTML]{3BAD6E}{RGB}}       & $6/10$  & $4/10$      & $3/10$        &    $1/10$ & $14/40$ \\
\textbf{\textcolor[HTML]{113ED9}{+Saliency}} & $8/10$  & $7/10$      & $6/10$        &    $6/10$ & $27/40$ \\
\bottomrule
\end{tabular}
\caption{Task success rates in real-world tabletop manipulation tasks for \textbf{\textcolor[HTML]{3BAD6E}{RGB}} and \textbf{\textcolor[HTML]{113ED9}{RGB+Saliency}} with MultiMAE.}
\vspace{-5px}
\label{tab:real_world_results}
\end{table}

\textbf{ViSaRL scales to real-robot tasks and is robust to distractor objects.} 
Videos of evaluation trajectories for each task can be found on the project website.
Table~\ref{tab:real_world_results} reports the task success rates on 10 evaluation rollouts.
Even on the easier \texttt{Pick Apple} task, using saliency augmented representations, \textbf{\textcolor[HTML]{113ED9}{RGB+Saliency}}, improves the success rate over \rgb.
On tasks with distractor objects and longer-horizon tasks such as \texttt{Put Apple in Bowl}, saliency-augmented representations nearly double the success rate. 

%% file: sections/conclusion.tex
\label{sec:conclusion}
We proposed to use human-annotated saliency as an additional input modality for solving challenging visual robot control tasks. 
We present a simple approach, ViSaRL, to utilize saliency to learn robust image representations enabling more sample-efficient and generalizable policy learning. 

\textbf{Limitations and Future Work.} 
One potential limitation of our user interface is that it could be tedious to collect saliency annotations when scaling to more complex real world applications or video saliency \cite{wang2018revisiting}. 
Future work could investigate alternative interfaces that will enable collecting more saliency data, e.g., area-based methods or by tracking the eye gaze of the user \cite{papadopoulos2014training}.

One can further evaluate the generalizability of ViSaRL on the recent benchmark, The Colosseum \cite{pumacay2024colosseum}, a suite of manipulation tasks design to measure the robustness of trained robot policies against visual perturbations. 

In this paper, we only considered static frame saliency maps for single-object manipulation tasks. 
We plan to extend our approach to handle longer-horizon multi-object tasks using video saliency models \cite{rudoy2013learning} which can learn to encode more flexible temporal saliency representations across a sequence of frames. 
This extension could be implemented by asking the human users to watch video clips of the trajectories and annotate saliency over these clips. 

%% file: sections/appendix/environment_description.tex
\textbf{Meta-World:} The observation space is a $64 \times 64 \times 3$ image. 
The action space $\mathcal{A}\subset \mathbb{R}^{4}$, is the $\Delta (xyz)$ of the end-effector, and a continuous scalar value for gripper torque. 
Object and goal positions are randomized at the start of every episode to prevent exploiting spurious correlation or memorizing trajectories to solve the task. 

\begin{figure}[h]
\centering
\centerline{\includegraphics[width=0.8\linewidth]{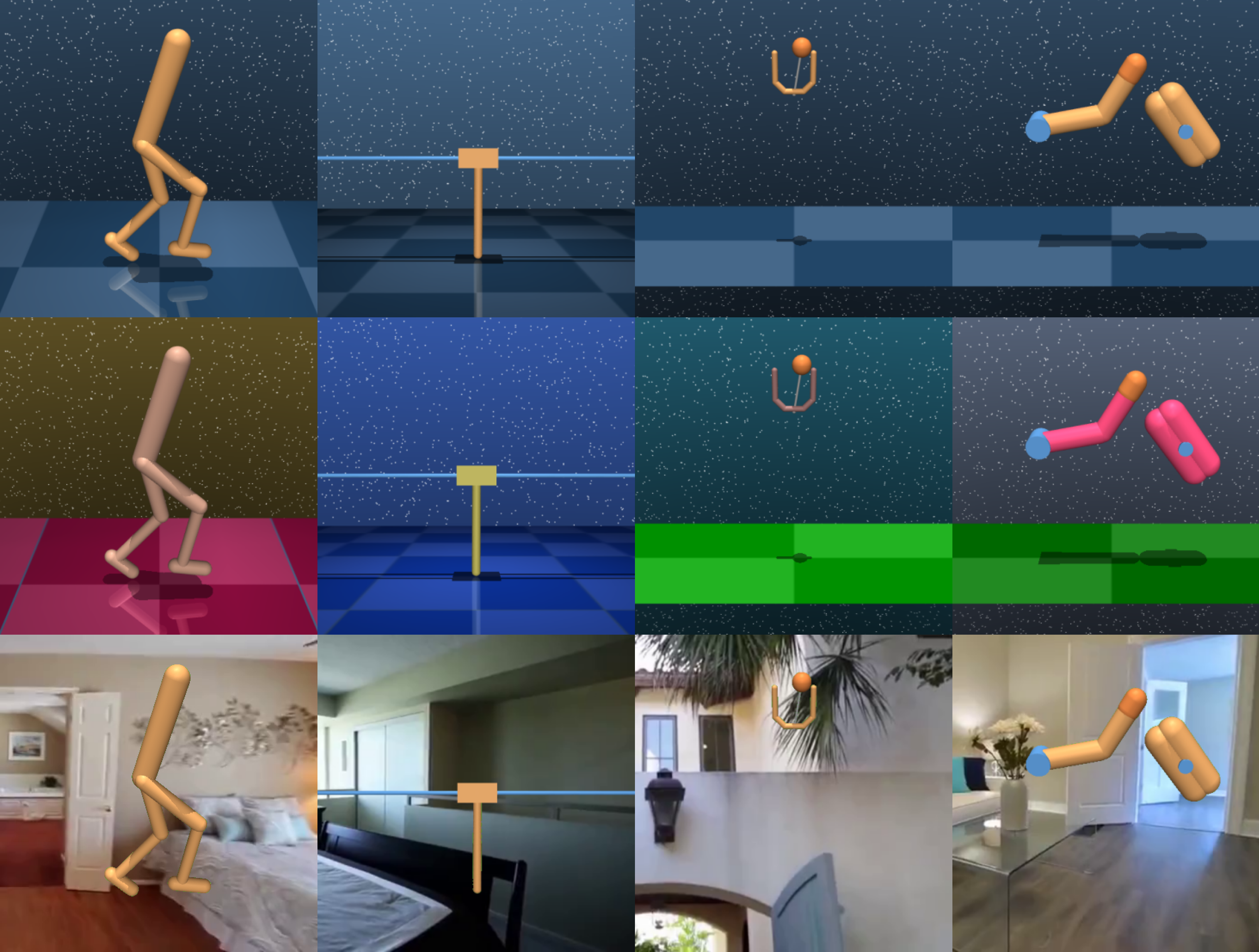}}
\caption{\textbf{DMControl Generalization Benchmark} (Top) Four continuous control tasks from the DMControl suite. (Middle and Bottom) We evaluate on the \texttt{color\_easy} and \texttt{video\_easy} settings to test the generalizability of the learned representations.}
\label{fig:dmc_tasks}
\end{figure}

\textbf{DMControl Generalization Benchmark} is built upon the DeepMind Control Suite, a widely used set of continuous control tasks designed to evaluate reinforcement learning algorithms. The observation space is $84 \times 84 \times 3$ RGB images. The action space is continuous and varies depending on the specific task, involving actions such as joint torques or target velocities for controlling the movement of robotic agents.

A key aspect of this benchmark is its emphasis on visual generalization. The DMControl Generalization Benchmark changing the background of the environment to either a static image or a video background. This setup challenges the agent’s ability to generalize, as it must learn image representations that are robust to visual distractions or changes in the background scene. 

The benchmark includes a range of tasks such as \texttt{Walker Walk}, \texttt{Cartpole Swing}, \texttt{Ball Catch}, and \texttt{Finger Spin}.
\texttt{Walker Walk} requires the agent to control a bipedal walker, coordinating its limbs to achieve stable locomotion and maintain balance while moving forward. 
In \texttt{Cartpole Swing}, the agent must swing up and stabilize a pole attached to a cart, balancing it in an upright position.
The \texttt{Ball Catch} task requires the agent to manipulate a robotic arm to successfully catch a falling ball. 
Lastly, \texttt{Finger Spin} involves spinning a small object around its axis using a dexterous robotic finger.

%% file: sections/appendix/hyperparameters.tex
\begin{table}[h]
\centering
\begin{tabular}{lc}
\toprule
\textbf{Hyperparameter}        & \textbf{Value}       \\ 
\midrule
Action repeat         & 1 \\
Discount factor $\gamma$            & 0.99       \\
Episode length         & 500        \\
Replay buffer size            & 1M        \\
Policy learning rate            & Adam(lr=1e-4, $\beta_{1}=0.9, \beta_{2}=0.999$)          \\
$\alpha$ learning rate            & Adam(lr=3e-4, $\beta_{1}=0.5, \beta_{2}=0.999$)  \\
Batch size         & 128       \\
Target network update              & 2         \\
Target network momentum $\tau$         & 0.05          \\
Environment steps & 1M \\
\bottomrule
\end{tabular}
\caption{\label{tab:hyperparameters} Hyperparameters used in SAC training.}
\end{table}

%% file: sections/appendix/implementation_details.tex
\label{sec:online_rl}
\textbf{Implementation details.} 
Table \ref{tab:hyperparameters} summarize the hyperparameters in ViSaRL and other baselines.
ViSaRL is agnostic to the choice of downstream RL algorithm.
For the CNN encoder experiments, we follow the implementation from \cite{laskin2020reinforcement, bertoin2022look} and train both the encoder and policy end-to-end using Soft-Actor Critic.
The encoder consist of a stack of 11 convolutional layers, each with 32 filters of 3 $\times$ 3 kernels, no padding, stride of 2 for the first and 1 for all others. 
This results in a feature map of dimension $32 \times 12 \times 12$ from an input image of shape $64 \times 64 \times 3$. 

The policy head $\pi_{\theta}$ and action-value functions $Q_{\phi_{i}}$ are parameterized by multi-layer perceptrons. 
The policy head is composed of a linear projection of dimension 100 with normalization followed by 3 linear layers with 1024 hidden units each and a final linear output layer for the action prediction. 
As the action spaces of DMControl suite and Meta-World are continuous, the policy outputs the mean and variance of a Gaussian distribution over actions. 

%% file: sections/appendix/real_robot_il.tex
\textbf{Real Robot Imitation Learning.} The downstream policy is trained using standard imitation learning. 
We train the saliency predictor and Transformer encoder using the same procedure outlined for the simulation experiments. 
Given our pretrained visual encoder, we form the state representation as a concatenation of the visual embedding and the robot proprioceptive information (e.g. joint positions). This yields a 271-dimensional state representation.

\begin{figure}[h]
\centering
\centerline{\includegraphics[width=\linewidth]{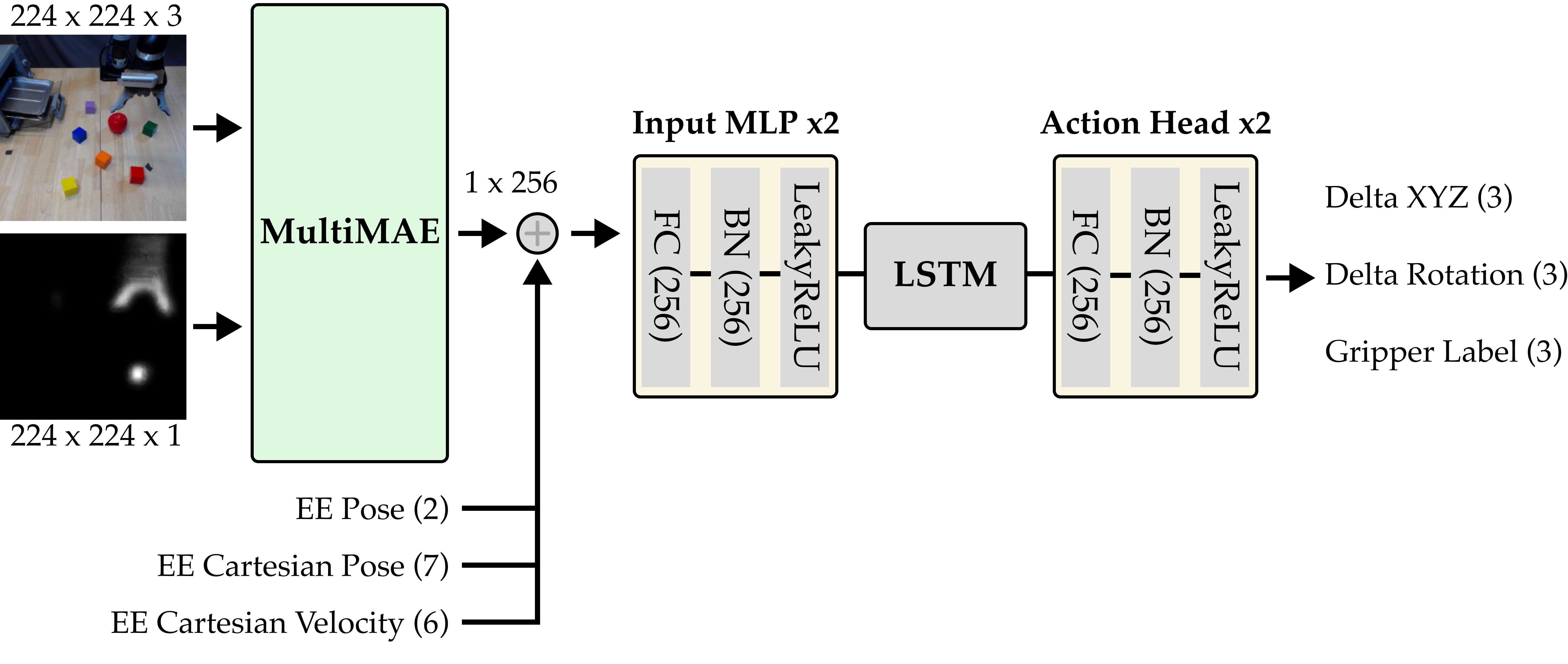}}
\caption{\textbf{Robot Policy Architecture} We use an LSTM policy with MLPs to embed the input and generate
continuous actions. 
The input is the cropped RGB image from an external camera
and the predicted saliency map. 
The visual input is contatenated with the proprioceptive information.}
\label{fig:robot_policy}
\end{figure}

\begin{table}[h]
\centering
\begin{tabular}{lc}
\toprule
\textbf{Hyperparameter}        & \textbf{Value}       \\ 
\midrule
Training epochs         & 500 \\
Subseq length ($H$)     & 15 \\ 
Batch size              & 64 \\
Control frequency       & 5 \\ 
Gripper weights         & [0.585, 0.08, 0.335] \\
MLP layers              & 2 \\
MLP hidden sizes        & [256, 256] \\
Normalization           & Batch \\
LSTM hidden size        & 256 \\    
\bottomrule
\end{tabular}
\caption{\label{tab:real_robot_hp} Hyperparameters used in the real robot experiment.}
\end{table}

\begin{figure*}[t]
\centering
\centerline{\includegraphics[width=\textwidth]{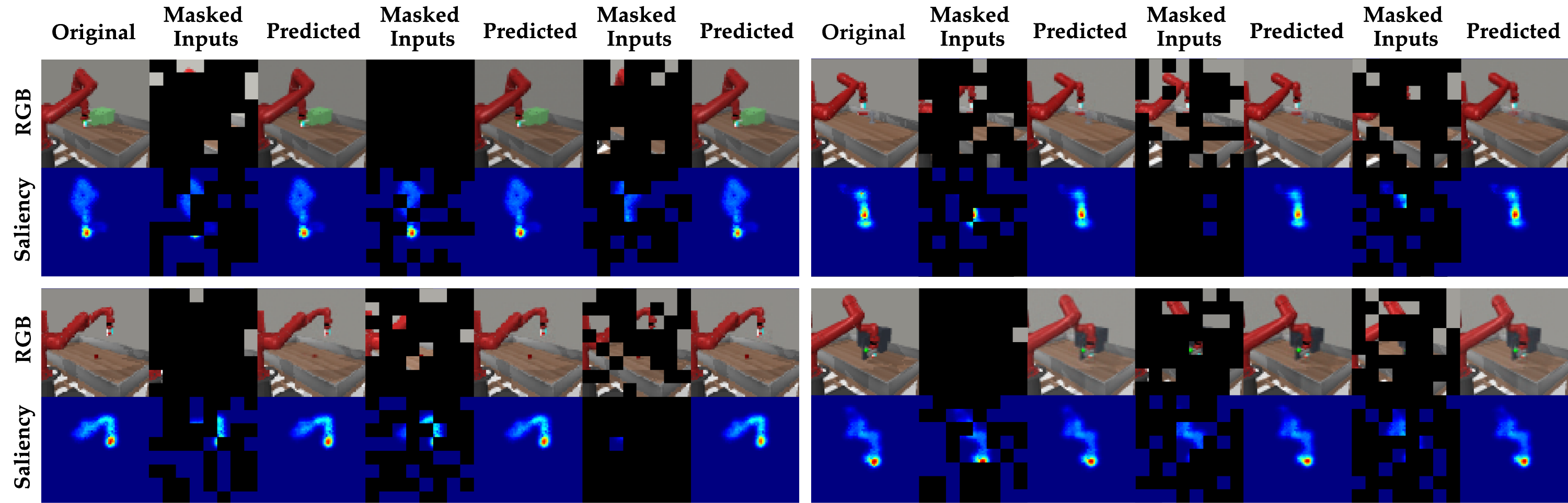}}
\caption{\textbf{MultiMAE predictions for different random masks.} We visualize the masked predictions for RGB observation from each of the four tasks. For each input image, we randomly sample three different masks from a uniform distribution between RGB and saliency. Only 1/4 of the total patches are unmasked. Even when there are a few unmasked patches from one modality, the reconstructions are still very accurate due to cross-modal interaction. Saliency maps are shown with color for the purposes of visualization.}
\label{fig:mask_vis}
\end{figure*}

The policy is implemented as an LSTM with 256-dimensional hidden states which autoregressively predicts the actions for the next $H$ timesteps where $H$ is a fixed window.
The state is processed by a 2-layer input MLP with hidden sizes [256, 256]
followed by a BatchNorm before being inputted to the LSTM.
The final hidden state of the LSTM is processed by a 2-layer MLP to predict a continuous action.
The full model architecture is shown in Figure \ref{fig:robot_policy}.
The action space, $ \mathcal{A} \in \mathbb{R}^{7}$, consists of $\Delta(x, y, z, \phi, \theta, \psi)$, and a continuous scalar value for gripper speed. 

%% file: sections/appendix/saliency_predictor.tex
\textbf{Saliency predictor training details.}
Since we are only using a small dataset for training the saliency predictor, we apply data augmentation to prevent overfitting and improve robustness of the model. After resizing the image to $224 \times 224$, we apply VerticalFlip and HorizontalFlip with a 50\% probability and ColorJitter(brightness=0.4, contrast=0.4, saturation=0.4, hue=0.2). We use AdamW optimizer with initial learning rate of 3e-4 and weight decay 0.005. We reduce the learning rate by 0.1 when loss does not decrease 0.001 with patience 100 and a minimum learning rate of 1e-6.

We opted to remove one local attention and one global attention decoder layer from the vanilla PiCANet, reducing our inference time per frame from 0.1 seconds to 0.01 seconds. 
Qualitatively, we observed little degradation in saliency prediction without these layers. 

\textbf{Choice of saliency predictor architecture.} 
Kummerer et al. \cite{kummerer2022deepgaze} introduces one of the current state-of-the-art saliency predictors: DeepGaze III. However, DeepGaze III is a scanpath model. A scanpath model predicts where a participant is likely to fixate given their fixation history. 
Since scanpath models leverage temporal information from human eye movement, it necessitates that the data collected be from an eye tracker. 
As our data is collected from clicks, it does not contain any temporal structure and as a result we cannot benefit from scanpath models. 

For future works, we plan to explore video saliency and how to exploit temporal patterns in human gaze to further improve the performance of our method.
Kummerer et al. \cite{kummerer2022deepgaze} also notes that scanpath models have worse performance than static models when the scene contains multiple small salient objects which is typically the case for tabletop manipulation.

\begin{figure}[h]
\begin{center}
\centerline{\includegraphics[width=0.5\textwidth]{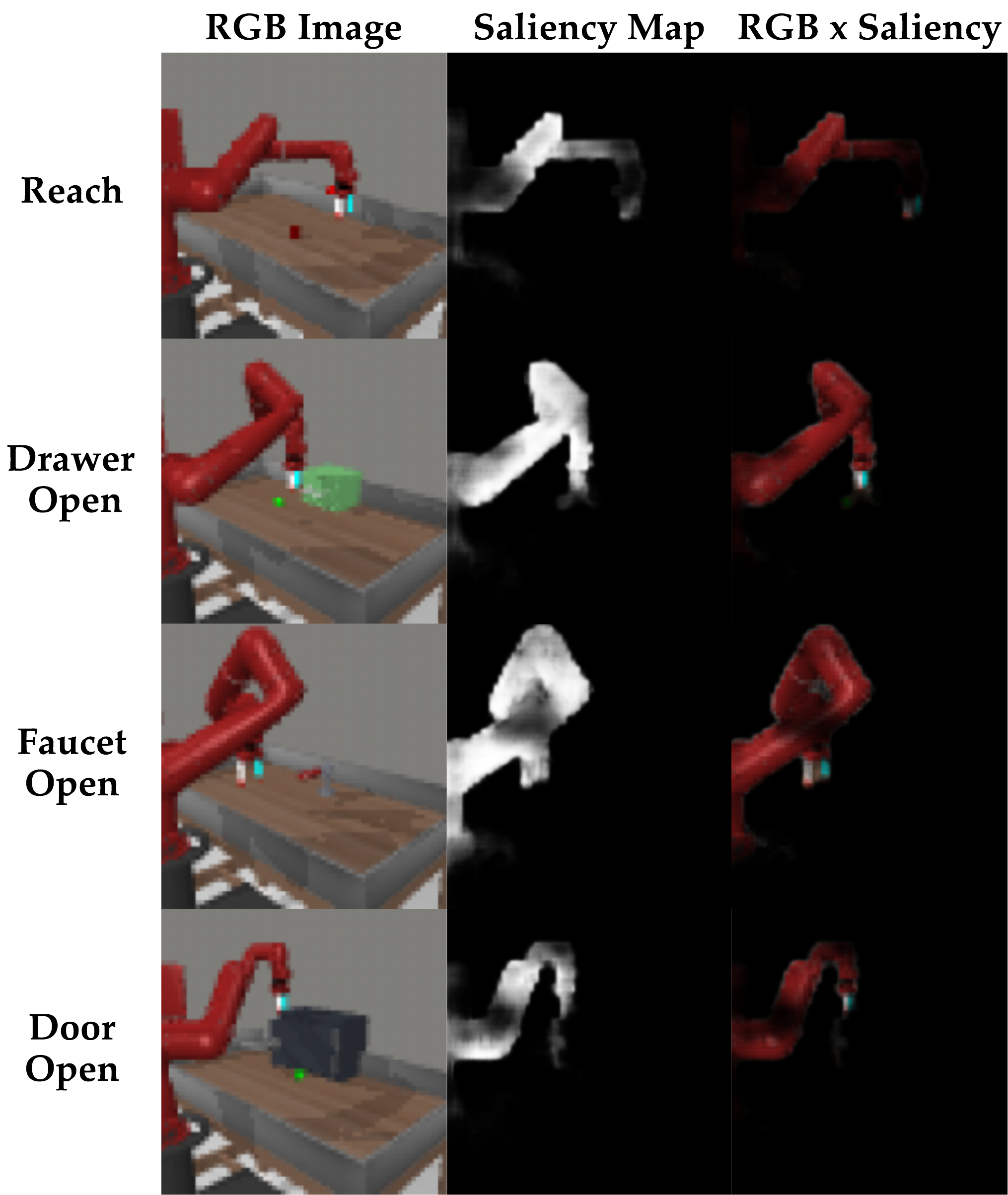}}
\caption{Zero-shot inference with pretrained PiCANet model for Meta-World observations. We observe that using only a pretrained saliency predictor will only identify the most prominent entity in the scene which in our case is the robot arm. However, task-relevant objects such as the block or the drawer, which are typically smaller in the image frame relative to the robot arm, are not highlighted.}
\label{fig:pretrained_saliency}
\end{center}
\end{figure}
We provide additional quantitative results in Table \ref{tab:compare_saliency_models} comparing the performance of different state-of-the-art saliency predictors including PiCANet 
~\cite{liu2018picanet}. 
We find PiCANet outperforms the two most recent DeepGaze models in both standard evaluation metrics.

\begin{table}[h]
\centering
\begin{tabular}{lcc}
\toprule
\textbf{Method}        & \textbf{Mean Absolute Error} & \textbf{F$_\zeta$ score}       \\ 
\midrule
DeepGaze II \cite{kummerer2016deepgaze}  & 0.0273 $\pm$ 0.006 & 0.7142 $\pm$ 0.021 \\
DeepGaze IIE \cite{linardos2021deepgaze} & 0.0153 $\pm$ 0.006 & 0.7283 $\pm$ 0.024 \\
PiCANet \cite{liu2018picanet} & 0.0032 $\pm$ 0.002 & 0.7970 $\pm$ 0.015 \\
\bottomrule
\end{tabular}
\caption{\label{tab:compare_saliency_models} Evaluation metrics measuring the prediction quality different state-of-the-art predictors.}
\end{table}

\textbf{Zero-shot saliency inference with pretrained PiCANet}. 
We experiment with zero-shot evaluation of a PiCANet saliency model pretrained on the training split of the DUTS dataset~\cite{wang2017learning}. 
This is currently one of the largest salient object detection datasets with 10,553 training images.

We observe that in a zero-shot setting, the pretrained model consistently segments out the robot arm (see Figure \ref{fig:pretrained_saliency}) even in different configurations, which is the most salient object in the scene.
However, it fails to identify the goal location or the drawer handle which are critical for completing the task. 
This demonstrates the need for human-annotated saliency which ViSaRL is able to capture.

\begin{table}[h]
\centering
\begin{tabular}{ccc}
\toprule
\textbf{Number of annotations}  & \textbf{Mean Absolute Error} & \textbf{F$_\zeta$ score}       \\ 
\midrule
5  & 0.0086 & 0.7983 \\
10 & 0.0063 & 0.7948 \\
20 & 0.0053 & 0.8006 \\
30 & 0.0020 & 0.8006 \\
\bottomrule
\end{tabular}
\caption{\label{tab:dataset_size} Effect of training dataset size on saliency prediction}
\end{table}

\textbf{Effect of number of annotated examples on saliency prediction}. 
We provide additional results in Table \ref{tab:dataset_size} investigating how the number of human annotations affects the performance and generalization of the learned saliency model.
We conduct an ablation over $x \in \{5, 10, 20, 30\}$ annotated examples. 
We observe that evaluation metrics improve with more training examples. 
Qualitatively, there is little difference between the predicted saliency maps using a model trained with 5 examples versus one trained with 30 examples. 
Even under different initial configurations, the model trained with only 5 annotated examples can generalize the saliency prediction to unseen environment states.

\begin{figure*}[h]
\begin{center}
\centerline{\includegraphics[width=\textwidth]{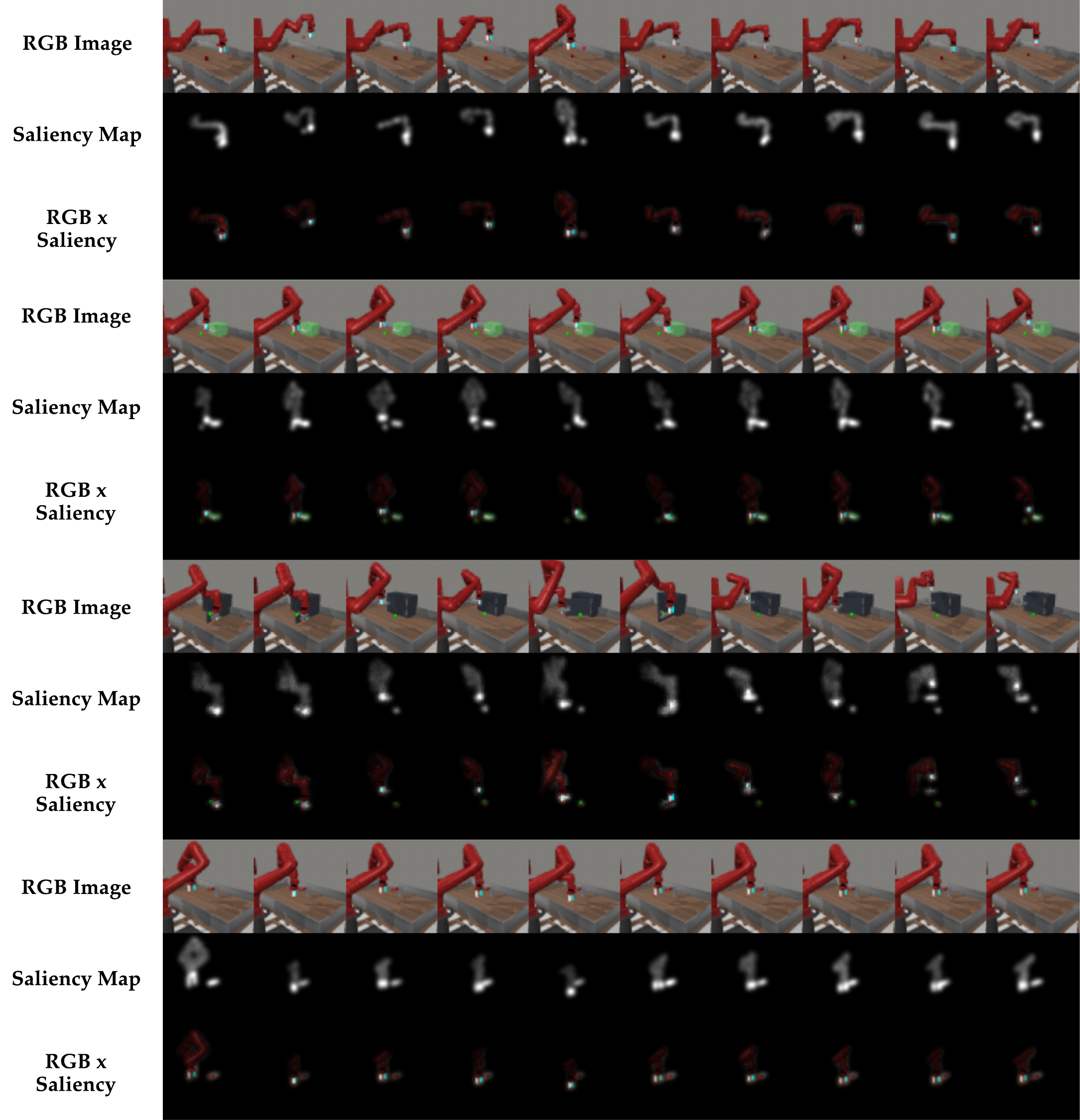}}
\caption{Visualization of randomly sampled environment observations, their corresponding saliency predicted by a pretrained saliency model, and masked RGB observation from the MetaWorld tasks.}
\label{fig:full_saliency_viz}
\end{center}
\end{figure*}

\begin{figure*}[h]
\begin{center}
\centerline{\includegraphics[width=\textwidth]{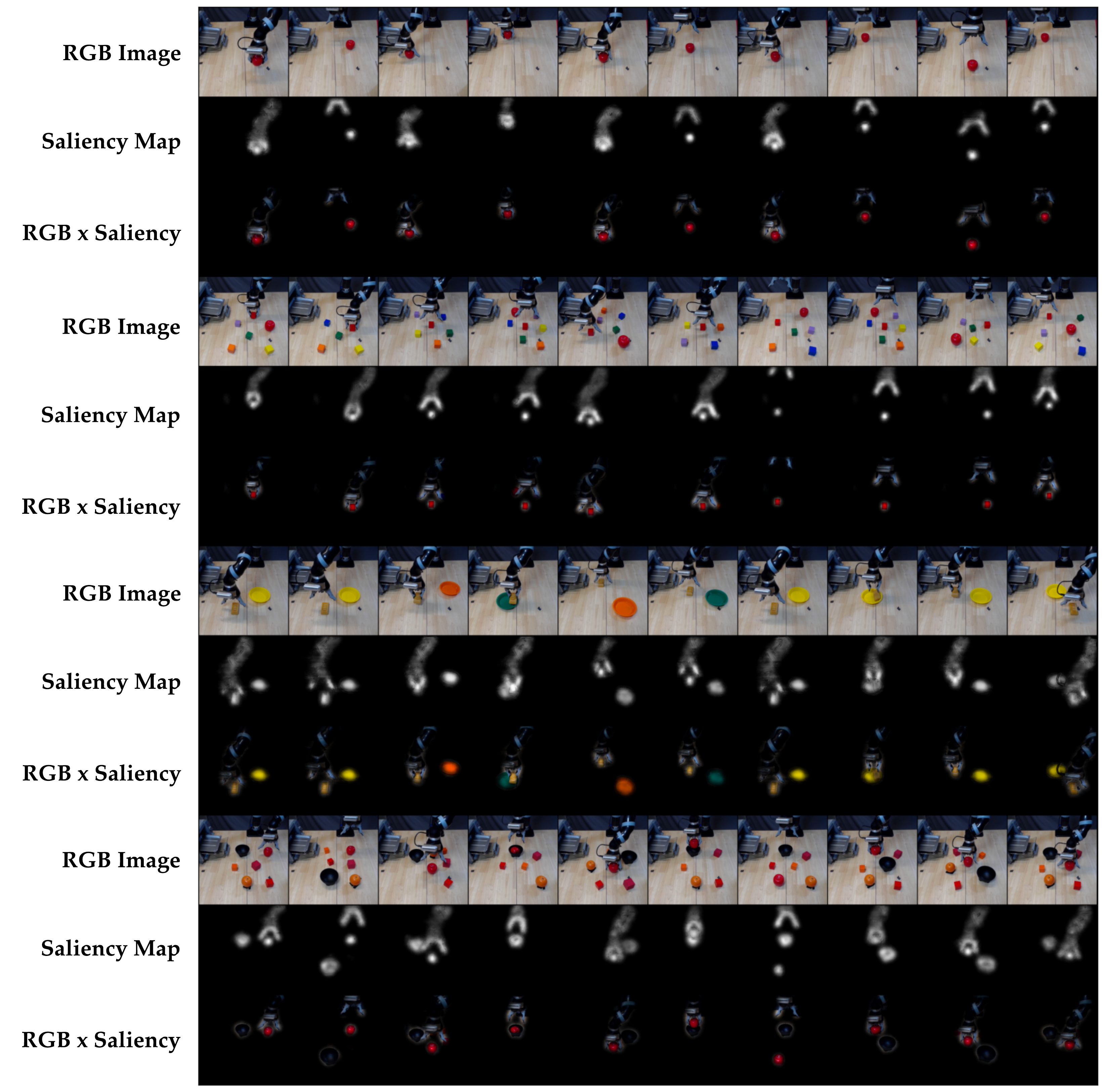}}
\caption{Visualization of randomly sampled environment observations, their corresponding saliency predicted by a pretrained saliency model, and masked RGB observation from real robot task demonstrations.}
\label{fig:full_saliency_viz_real}
\end{center}
\end{figure*}

%% file: root.bbl
\begin{thebibliography}{10}
\providecommand{\url}[1]{#1}
\csname url@rmstyle\endcsname
\providecommand{\newblock}{\relax}
\providecommand{\bibinfo}[2]{#2}
\providecommand\BIBentrySTDinterwordspacing{\spaceskip=0pt\relax}
\providecommand\BIBentryALTinterwordstretchfactor{4}
\providecommand\BIBentryALTinterwordspacing{\spaceskip=\fontdimen2\font plus
\BIBentryALTinterwordstretchfactor\fontdimen3\font minus \fontdimen4\font\relax}
\providecommand\BIBforeignlanguage[2]{{%
\expandafter\ifx\csname l@#1\endcsname\relax
\typeout{** WARNING: IEEEtran.bst: No hyphenation pattern has been}%
\typeout{** loaded for the language `#1'. Using the pattern for}%
\typeout{** the default language instead.}%
\else
\language=\csname l@#1\endcsname
\fi
#2}}

\bibitem{darby2021development}
K.~P. Darby, S.~W. Deng, D.~B. Walther, and V.~M. Sloutsky, ``{The development of attention to objects and scenes: From object-biased to unbiased},'' \emph{Child development}, 2021.

\bibitem{itti1998model}
L.~Itti, C.~Koch, and E.~Niebur, ``{A model of saliency-based visual attention for rapid scene analysis},'' \emph{IEEE Transactions on Pattern Analysis and Machine Intelligence}, 1998.

\bibitem{laskin2020curl}
M.~Laskin, A.~Srinivas, and P.~Abbeel, ``{CURL: Contrastive Unsupervised Representations for Reinforcement Learning},'' in \emph{International Conference on Machine Learning (ICML)}, 2020.

\bibitem{laskin2020reinforcement}
M.~Laskin, K.~Lee, A.~Stooke, L.~Pinto, P.~Abbeel, and A.~Srinivas, ``{Reinforcement Learning with Augmented Data},'' \emph{Neural Information Processing Systems (NeurIPS)}, 2020.

\bibitem{tunyasuvunakool2020dm_control}
S.~Tunyasuvunakool, A.~Muldal, Y.~Doron, S.~Liu, S.~Bohez, J.~Merel, T.~Erez, T.~Lillicrap, N.~Heess, and Y.~Tassa, ``{dm\_control: Software and Tasks for Continuous Control},'' \emph{Software Impacts}, 2020.

\bibitem{yu2020meta}
T.~Yu, D.~Quillen, Z.~He, R.~Julian, K.~Hausman, C.~Finn, and S.~Levine, ``{Meta-World: A Benchmark and Evaluation for Multi-Task and Meta Reinforcement Learning},'' in \emph{Conference on Robot Learning (CoRL)}, 2020.

\bibitem{cabi2019scaling}
S.~Cabi, S.~G. Colmenarejo, A.~Novikov, K.~Konyushkova, S.~Reed, R.~Jeong, K.~Zolna, Y.~Aytar, D.~Budden, M.~Vecerik, \emph{et~al.}, ``{Scaling data-driven robotics with reward sketching and batch reinforcement learning},'' \emph{Robotics: Science and Systems (RSS)}, 2020.

\bibitem{bobu2021feature}
A.~Bobu, M.~Wiggert, C.~Tomlin, and A.~D. Dragan, ``{Feature Expansive Reward Learning: Rethinking Human Input},'' in \emph{Human-Robot Interaction (HRI)}, 2021.

\bibitem{wilde2022learning}
N.~Wilde, E.~Biyik, D.~Sadigh, and S.~L. Smith, ``{Learning Reward Functions from Scale Feedback},'' in \emph{Conference on Robot Learning (CoRL)}, 2022.

\bibitem{tao2022executable}
S.~Tao, X.~Li, T.~Mu, Z.~Huang, Y.~Qin, and H.~Su, ``{Abstract-to-Executable Trajectory Translation for One-Shot Task Generalization},'' in \emph{Neural Information Processing Systems (NeurIPS) Deep Reinforcement Learning Workshop}, 2022.

\bibitem{tong2010full}
Y.~Tong, H.~Konik, F.~Cheikh, and A.~Tremeau, ``{Full Reference Image Quality Assessment Based on Saliency Map Analysis},'' \emph{Journal of Imaging Science and Technology}, 2010.

\bibitem{wang2016beyond}
X.~Wang, L.~Gao, J.~Song, and H.~Shen, ``{Beyond Frame-level CNN: Saliency-Aware 3-D CNN With LSTM for Video Action Recognition},'' \emph{IEEE Signal Processing Letters}, 2016.

\bibitem{das2017human}
A.~Das, H.~Agrawal, L.~Zitnick, D.~Parikh, and D.~Batra, ``{Human Attention in Visual Question Answering: Do Humans and Deep Networks Look at the Same Regions?}'' \emph{Computer Vision and Image Understanding}, 2017.

\bibitem{li2011saliency}
Q.~Li, Y.~Zhou, and J.~Yang, ``{Saliency Based Image Segmentation},'' in \emph{International Conference on Information and Multimedia Technology (ICIMT)}, 2011.

\bibitem{simonyan2013deep}
K.~Simonyan, A.~Vedaldi, and A.~Zisserman, ``{Deep Inside Convolutional Networks: Visualising Image Classification Models and Saliency Maps},'' \emph{arXiv preprint arXiv:1312.6034}, 2013.

\bibitem{mundhenk2019efficient}
T.~N. Mundhenk, B.~Y. Chen, and G.~Friedland, ``{Efficient Saliency Maps for Explainable AI},'' \emph{arXiv preprint arXiv:1911.11293}, 2019.

\bibitem{zhao2016person}
R.~Zhao, W.~Oyang, and X.~Wang, ``{Person Re-Identification by Saliency Learning},'' \emph{IEEE Transactions on Pattern Analysis and Machine Intelligence}, 2016.

\bibitem{atrey2019exploratory}
A.~Atrey, K.~Clary, and D.~Jensen, ``{Exploratory Not Explanatory: Counterfactual Analysis of Saliency Maps for Deep Reinforcement Learning},'' \emph{International Conference on Learning Representations (ICLR)}, 2020.

\bibitem{rosynski2020gradient}
M.~Rosynski, F.~Kirchner, and M.~Valdenegro-Toro, ``{Are Gradient-based Saliency Maps Useful in Deep Reinforcement Learning?}'' \emph{arXiv preprint arXiv:2012.01281}, 2020.

\bibitem{boyd2023cyborg}
A.~Boyd, P.~Tinsley, K.~W. Bowyer, and A.~Czajka, ``{CYBORG: Blending Human Saliency Into the Loss Improves Deep Learning},'' in \emph{Winter Conference on Applications of Computer Vision (WACV)}, 2023.

\bibitem{bertoin2022look}
D.~Bertoin, A.~Zouitine, M.~Zouitine, and E.~Rachelson, ``{Look where you look! Saliency-guided Q-networks for generalization in visual Reinforcement Learning},'' \emph{Neural Information Processing Systems (NeurIPS)}, 2022.

\bibitem{achanta2012slic}
R.~Achanta, A.~Shaji, K.~Smith, A.~Lucchi, P.~Fua, and S.~S{\"u}sstrunk, ``{SLIC Superpixels Compared to State-of-the-Art Superpixel Methods},'' \emph{IEEE Transactions on Pattern Analysis and Machine Intelligence}, 2012.

\bibitem{boyd2022human}
A.~Boyd, K.~W. Bowyer, and A.~Czajka, ``{Human-Aided Saliency Maps Improve Generalization of Deep Learning},'' in \emph{Winter Conference on Applications of Computer Vision (WACV)}, 2022.

\bibitem{nair2022r3m}
S.~Nair, A.~Rajeswaran, V.~Kumar, C.~Finn, and A.~Gupta, ``{R3M: A Universal Visual Representation for Robot Manipulation},'' \emph{Conference on Robot Learning (CoRL)}, 2022.

\bibitem{karamcheti2023language}
S.~Karamcheti, S.~Nair, A.~S. Chen, T.~Kollar, C.~Finn, D.~Sadigh, and P.~Liang, ``{Language-Driven Representation Learning for Robotics},'' \emph{arXiv preprint arXiv:2302.12766}, 2023.

\bibitem{sax2018mid}
A.~Sax, B.~Emi, A.~R. Zamir, L.~Guibas, S.~Savarese, and J.~Malik, ``{Mid-Level Visual Representations Improve Generalization and Sample Efficiency for Learning Visuomotor Policies},'' \emph{Conference on Robot Learning (CoRL)}, 2019.

\bibitem{bachmann2022multimae}
R.~Bachmann, D.~Mizrahi, A.~Atanov, and A.~Zamir, ``{MultiMAE: Multi-modal Multi-task Masked Autoencoders},'' \emph{European Conference on Computer Vision (ECCV)}, 2022.

\bibitem{liu2018picanet}
N.~Liu, J.~Han, and M.-H. Yang, ``{PiCANet: Learning Pixel-wise Contextual Attention for Saliency Detection},'' in \emph{Computer Vision and Pattern Recognition (CVPR)}, 2018.

\bibitem{he2022masked}
K.~He, X.~Chen, S.~Xie, Y.~Li, P.~Doll{\'a}r, and R.~Girshick, ``{Masked Autoencoders Are Scalable Vision Learners},'' in \emph{Computer Vision and Pattern Recognition (CVPR)}, 2022.

\bibitem{dosovitskiy2020image}
A.~Dosovitskiy, L.~Beyer, A.~Kolesnikov, D.~Weissenborn, X.~Zhai, T.~Unterthiner, M.~Dehghani, M.~Minderer, G.~Heigold, S.~Gelly, \emph{et~al.}, ``{An Image is Worth 16x16 Words: Transformers for Image Recognition at Scale},'' \emph{International Conference on Learning Representations (ICLR)}, 2021.

\bibitem{haarnoja2018soft}
T.~Haarnoja, A.~Zhou, K.~Hartikainen, G.~Tucker, S.~Ha, J.~Tan, V.~Kumar, H.~Zhu, A.~Gupta, P.~Abbeel, \emph{et~al.}, ``{Soft Actor-Critic Algorithms and Applications},'' \emph{International Conference on Machine Learning (ICML)}, 2018.

\bibitem{hansen2021generalization}
N.~Hansen and X.~Wang, ``Generalization in reinforcement learning by soft data augmentation,'' in \emph{2021 IEEE International Conference on Robotics and Automation (ICRA)}.\hskip 1em plus 0.5em minus 0.4em\relax IEEE, 2021, pp. 13\,611--13\,617.

\bibitem{wang2018revisiting}
W.~Wang, J.~Shen, F.~Guo, M.-M. Cheng, and A.~Borji, ``{Revisiting Video Saliency: A Large-scale Benchmark and a New Model},'' in \emph{Computer Vision and Pattern Recognition (CVPR)}, 2018.

\bibitem{papadopoulos2014training}
D.~P. Papadopoulos, A.~D. Clarke, F.~Keller, and V.~Ferrari, ``{Training Object Class Detectors from Eye Tracking Data},'' in \emph{European Conference on Computer Vision (ECCV)}, 2014.

\bibitem{pumacay2024colosseum}
W.~Pumacay, I.~Singh, J.~Duan, R.~Krishna, J.~Thomason, and D.~Fox, ``The colosseum: A benchmark for evaluating generalization for robotic manipulation,'' \emph{arXiv preprint arXiv:2402.08191}, 2024.

\bibitem{rudoy2013learning}
D.~Rudoy, D.~B. Goldman, E.~Shechtman, and L.~Zelnik-Manor, ``{Learning video saliency from human gaze using candidate selection},'' in \emph{Computer Vision and Pattern Recognition (CVPR)}, 2013.

\bibitem{kummerer2022deepgaze}
M.~K{\"u}mmerer, M.~Bethge, and T.~S. Wallis, ``{Deepgaze iii: Modeling free-viewing human scanpaths with deep learning},'' \emph{Journal of Vision}, 2022.

\bibitem{kummerer2016deepgaze}
M.~K{\"u}mmerer, T.~S. Wallis, and M.~Bethge, ``{DeepGaze II: Reading fixations from deep features trained on object recognition},'' \emph{arXiv preprint arXiv:1610.01563}, 2016.

\bibitem{linardos2021deepgaze}
A.~Linardos, M.~K{\"u}mmerer, O.~Press, and M.~Bethge, ``{DeepGaze IIE: Calibrated prediction in and out-of-domain for state-of-the-art saliency modeling},'' in \emph{International Conference on Computer Vision (ICCV)}, 2021.

\bibitem{wang2017learning}
L.~Wang, H.~Lu, Y.~Wang, M.~Feng, D.~Wang, B.~Yin, and X.~Ruan, ``{Learning to Detect Salient Objects with Image-level Supervision},'' in \emph{Computer Vision and Pattern Recognition (CVPR)}, 2017.

\end{thebibliography}
